\documentclass[a4paper]{styles/svproc}
\usepackage{url}

\usepackage{graphicx}

\usepackage{multirow}
\usepackage[normalem]{ulem}
\usepackage{subcaption}
\usepackage{hyperref}

\usepackage{times} 
\usepackage{amsmath} 
\usepackage{amssymb}  
\usepackage{outlines}
\usepackage{float}
\usepackage{makebox}
\usepackage{algorithm}
\usepackage{algorithmic}
\usepackage{array}
\usepackage{indentfirst}
\newcolumntype{P}[1]{>{\centering\arraybackslash}p{#1}} 
\newcolumntype{M}[1]{>{\centering\arraybackslash}m{#1}}

\useunder{\uline}{\ul}{}

\newcommand{\gripper}{$w_g$}
\newcommand{\deltaZ}{$\Delta z$}
\newcommand{\finalZ}{$z_f$}

\newcommand{\playaudio}{$\mathcal{A}_{play}$}
\newcommand{\cutaudio}{$\mathcal{A}_{cut}$}
\newcommand{\proprio}{$\mathcal{P}$}

\newcommand{\foodlabel}{$\mathcal{F}$}
\newcommand{\slicelabel}{$\mathcal{S}$}

\begin{document}
\mainmatter              
%

\title{Playing with Food: Learning Food Item Representations through Interactive Exploration}

\titlerunning{Playing with Food}  
%
\author{Amrita Sawhney\footnote{Equal Contribution \\ This work was supported by Sony AI.} \and Steven Lee\footnotemark[\value{footnote}] \and Kevin Zhang\footnotemark[\value{footnote}] \and \\ Manuela Veloso \and
 Oliver Kroemer}
%
\authorrunning{Sawhney, Lee, Zhang, Veloso, and Kroemer} 
%
\tocauthor{Amrita Sawhney, Steven Lee, Kevin Zhang, Manuela Veloso, and
Oliver Kroemer}
\institute{Carnegie Mellon University, Pittsburgh PA 15232, USA\\
\email{\{amritasa, stevenl3, klz1, mmv, okroemer\}@cs.cmu.edu}}

\maketitle              

\begin{abstract}
A key challenge in robotic food manipulation is modeling the material properties of diverse and deformable food items.
We propose using a multimodal sensory approach to interact and play with food that facilitates the ability to distinguish these properties across food items.
First, we use a robotic arm and an array of sensors, which are synchronized using ROS, to collect a diverse dataset consisting of 21 unique food items with varying slices and properties.
Afterwards, we learn visual embedding networks that utilize a combination of proprioceptive, audio, and visual data to encode similarities among food items using a triplet loss formulation. 
Our evaluations show that embeddings learned through interactions can successfully increase performance in a wide range of material and shape classification tasks.
We envision that these learned embeddings can be utilized as a basis for planning and selecting optimal parameters for more material-aware robotic food manipulation skills.
Furthermore, we hope to stimulate further innovations in the field of food robotics by sharing this food playing dataset with the research community.
\end{abstract}

\section{Introduction}

Knowledge of an object's material properties is important for robots learning to perform tasks that involve physical interactions.
However, obtaining material properties for deformable objects can be difficult and time consuming. 
Food items, in particular, vary widely between and within food types depending on factors such as how they were grown, how they were stored, and whether they have been cooked \cite{figura2007food}. 
As a result of our prior knowledge, humans typically can ascertain the basic properties of many different food items using only vision.
For properties that are not always clearly conveyed through vision, humans will touch or interact with objects in order to disambiguate its internal material properties, such as knocking on a watermelon to determine its ripeness. 
Analogously, we believe that the ability to distinguish properties between food items can be learned using multimodal sensor data from interactive robot exploration~\cite{kroemer2019review}. 

In this paper, we present a unique multimodal food interaction dataset consisting of vision, audio, proprioceptive, and force data acquired autonomously through robot interactions with a variety of food items.
Additionally, we use this dataset to explore a self-supervised method for learning embeddings that encode various food material properties and use visual data as input. 
The network used to learn these embeddings is trained using a triplet loss formulation, which groups similar and dissimilar samples based on the different types of interactive sensor data. 
In this manner, the robot can learn complex representations of these items autonomously without the need for subjective and time-consuming human labels.

To demonstrate the utility of the learned representations, we subsequently use the learned embeddings as input features to train regressors and classifiers for different tasks and compare their performances with baseline vision-only and audio-only approaches. 
Our experiments show that regressors and classifiers that use our learned embeddings outperform similar baseline networks on a variety of tasks, indicating that the visual embedding network encodes additional material property information from the different modalities, without requiring the robot to interact with the object at test time.
Our project website is located \href{https://sites.google.com/view/playing-with-food}{here}\footnote{\url{https://sites.google.com/view/playing-with-food}} along with a link to our dataset \href{https://tinyurl.com/playing-with-food-dataset}{here}\footnote{\url{https://tinyurl.com/playing-with-food-dataset}}.




\section{Related Work}

Many recent works have utilized simulations in order to learn dynamics models and material properties of deformable objects \cite{matas2018simtoreal,wu2020learning}. 
For example, Matl et al.~\cite{matl2020inferring} collect data on granular materials and compare the visual depth information with simulation results in order to infer their properties. 
Yan et al.~\cite{yan2020learning} learn latent dynamics models and visual representations of deformable objects by manipulating them in simulation and using contrastive estimation, which is similar to our approach.
In comparison, we worked with real-world robots and objects to collect data since representations of food that are learned through a simulation environment may not accurately transfer to real world due to variable behaviors during complex tasks, such as large plastic deformations during cutting.
There are simulators that can simulate large elasto-plastic deformation~\cite{chentanez2016real}, but they are computationally expensive, unavailable to the public, and have not yet shown their efficacy in this particular domain. 

Other works also use a variety of multimodal sensors to inform a robot of deformable object properties in order to better manipulate them~\cite{boonvisut2012estimation,jia2018manipulating,li2020review}.
Erickson et al.~\cite{erickson2020multimodal} use a highly specialized near-infrared spectrometer and texture imaging to classify the materials of objects.
Meanwhile, Feng et al.~\cite{feng2019robot} use a visual network and forces to determine the best location at which to skewer a variety of food items for bite acquisition. 
Finally, Zhang et al.~\cite{zhang2019leveraging} use forces and contact microphones to classify the hardness of ingredients in order to adjust the cutting parameters for slicing actions.
In contrast to these works, we only use overhead images as input to our embedding network during inference time, while still incorporating multi-modal data during training.

Numerous works have focused on using interactive perception to learn about objects in the environment~\cite{bohg2017interactive}.
Katz and Brock~\cite{katz2008manipulating} used interactive perception to determine the location and type of articulation on a variety of random objects.
Sinapov et al. \cite{sinapov2011interactive,sinapov2014grounding,tatiya2019deep} had a robot interact with objects using vision, proprioception, and audio in order to categorize them.
Chu et al.~\cite{chu2013using} utilized 2 Biotac sensors on a PR2 along with 5 exploratory actions in order to learn human labeled adjectives from haptic feedback.
In our work, we focus our exploratory data collection on deformable food objects instead of rigid objects.
Unlike the Epic Kitchens dataset \cite{damen2020epic}, which captures a first person view of humans cooking, we capture proprioceptive information using a robot and repeat consistent interactive actions across food items. 

Finally, researchers have been interested in learning deep embeddings for objects in order to create simpler representations for a variety of tasks~\cite{bengio2013representation,xie2016unsupervised}; however, performing the same task for food items has not been studied extensively. 
Sharma et al.~\cite{sharma2019learning} learn a semantic embedding network to represent the size of food slices in order to plan a sequence of cuts.
On the other hand, Isola et al.~\cite{StatesAndTransformations} used human labeled adjectives to generalize the transformation of object states, such as the ripeness in fruit, over time.
Although the above works learn embeddings for food objects, both of them focus on using solely visual inputs during both train and test time, while we incorporate additional synchronized multi-modal sensory information during the training of our vision-based embedding networks.

\section{Robotic Food Manipulation Dataset}

\begin{figure}[t]
    \centerline{\includegraphics[width=0.9\textwidth]{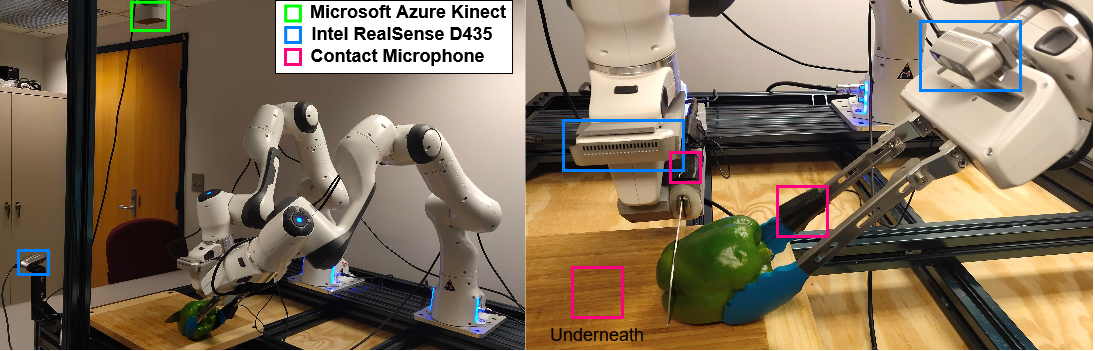}}
    \caption{Our cutting experimental setup.}
    \label{fig:cutting_robot_setup}
\end{figure}

\begin{figure}[t]
    \centerline{\includegraphics[width=\textwidth]{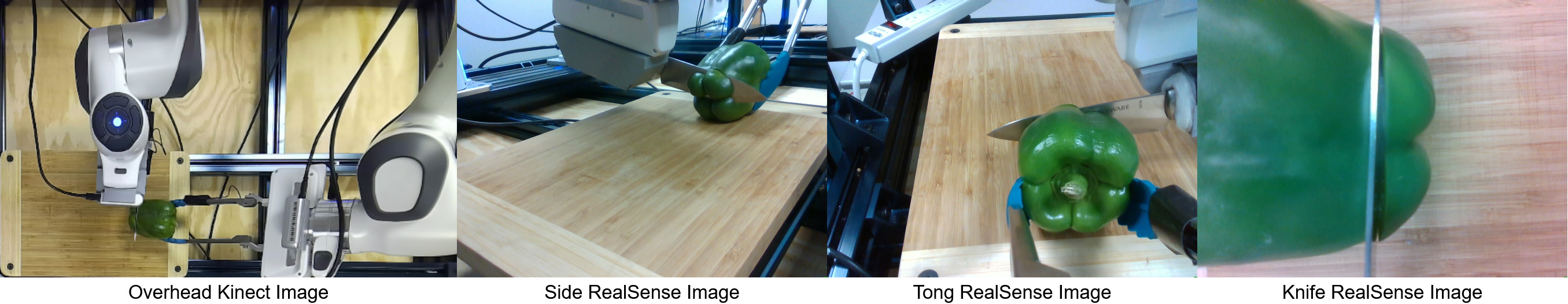}}
    \caption{Images from the 4 cameras spread around our cutting experimental setup.}
    \label{fig:cutting_robot_images}
    \vspace{-1em}
\end{figure}

\subsection{Experimental Setup}
\label{sec:exp_setup}

Our data collection pipeline involves two different experimental setups: one for robotic food cutting and another for food playing, wherein the robot interacts with the food slices that have already been cut. We collected multi-modal sensor data during the cutting and playing data collection processes using our Franka robot control framework \cite{zhang2020modular}. The sections below detail the setups for both cutting and playing in more detail.

\subsubsection{Robot Cutting:}
Our experimental setup for cutting data collection consists of two Franka Emika Panda Arms mounted on a Vention frame with a cutting board in the center, as shown in Fig.\ref{fig:cutting_robot_setup}. 
One arm is grasping a custom knife attachment while the other has a set of 8" kitchen tongs mounted to its fingers.
There are four cameras attached to the setup: an overhead Microsoft Azure Kinect Camera, a side-view Intel Realsense D435, another D435 mounted above the wrist of the robot holding the knife, and a third D435 mounted on the wrist of the robot holding the tongs. Sample images from each camera are shown in Fig. \ref{fig:cutting_robot_images}.
In addition, there are 3 contact microphones: one mounted underneath the center of the cutting board, another mounted on the knife attachment, and the last mounted on the tongs.

\begin{figure}[t]
    \centerline{\includegraphics[width=0.65\textwidth]{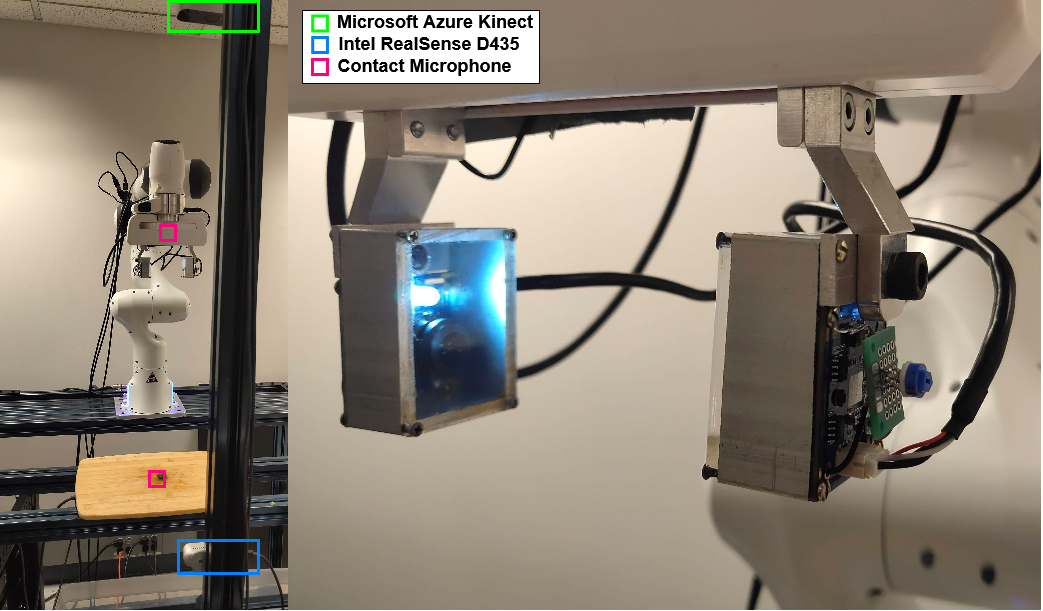}}
    \caption{Our playing experimental setup.}
    \label{fig:playing_robot_setup}
\end{figure}

\begin{figure}[t]
    \centerline{\includegraphics[width=0.95\textwidth]{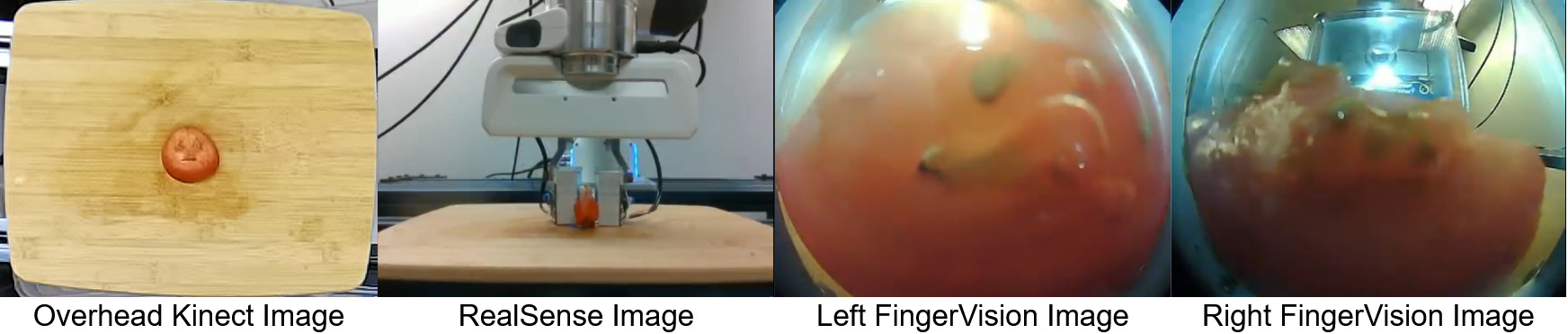}}
    \caption{Examples of Kinect, Realsense, and FingerVision images of a cut tomato.}
    \label{fig:playing_robot_images}
    \vspace{-1em}
\end{figure}

\vspace{-1em}
\subsubsection{Robot Playing:}
For our playing setup, we have a single Franka Emika Panda Arm mounted on a Vention frame with a cutting board in the center, as shown in Fig.~\ref{fig:playing_robot_setup}. 
We mounted an overhead Microsoft Azure Kinect Camera and a frontal Intel Realsense D435 that faces the robot. 
We attach a fisheye 1080P USB Camera\footnote{\url{https://www.amazon.com/180degree-Fisheye-Camera-usb-Android-Windows/dp/B00LQ854AG/}} to each fingertip as in FingerVision \cite{yamaguchi2016combining}, except we use a laser-cut clear acrylic plate cover instead of a soft gel-based cover over the camera.
This acrylic plate allows us to observe the compression of the object being grasped relative to fingertips.
We also added a white LED to better illuminate the object while it is being grasped.
Images from all of the cameras are shown in Fig. \ref{fig:playing_robot_images}.
We have 2 contact microphones on the setup: one mounted underneath the center of the cutting board and the other mounted on the back of the Franka Panda hand. 
The Piezo contact microphones\footnote{\url{https://www.amazon.com/Agile-Shop-Contact-Microphone-Pickup-Guitar/dp/B07HVFTGTH/}} from both setups capture vibro-tactile feedback through the cutting board, fingers, and tools.
The audio from the contact microphones of both the cutting and playing setup are captured using a Behringer UMC404HD Audio Interface\footnote{\url{https://www.amazon.com/BEHRINGER-Audio-Interface-4-Channel-UMC404HD/dp/B00QHURLHM}} and synchronized with ROS using sounddevice\_ros \cite{sounddevice_ros}.

\subsection{Data Collection}
\label{data_col}

\begin{figure}[t]
    \centering
    \includegraphics[width=0.9\textwidth]{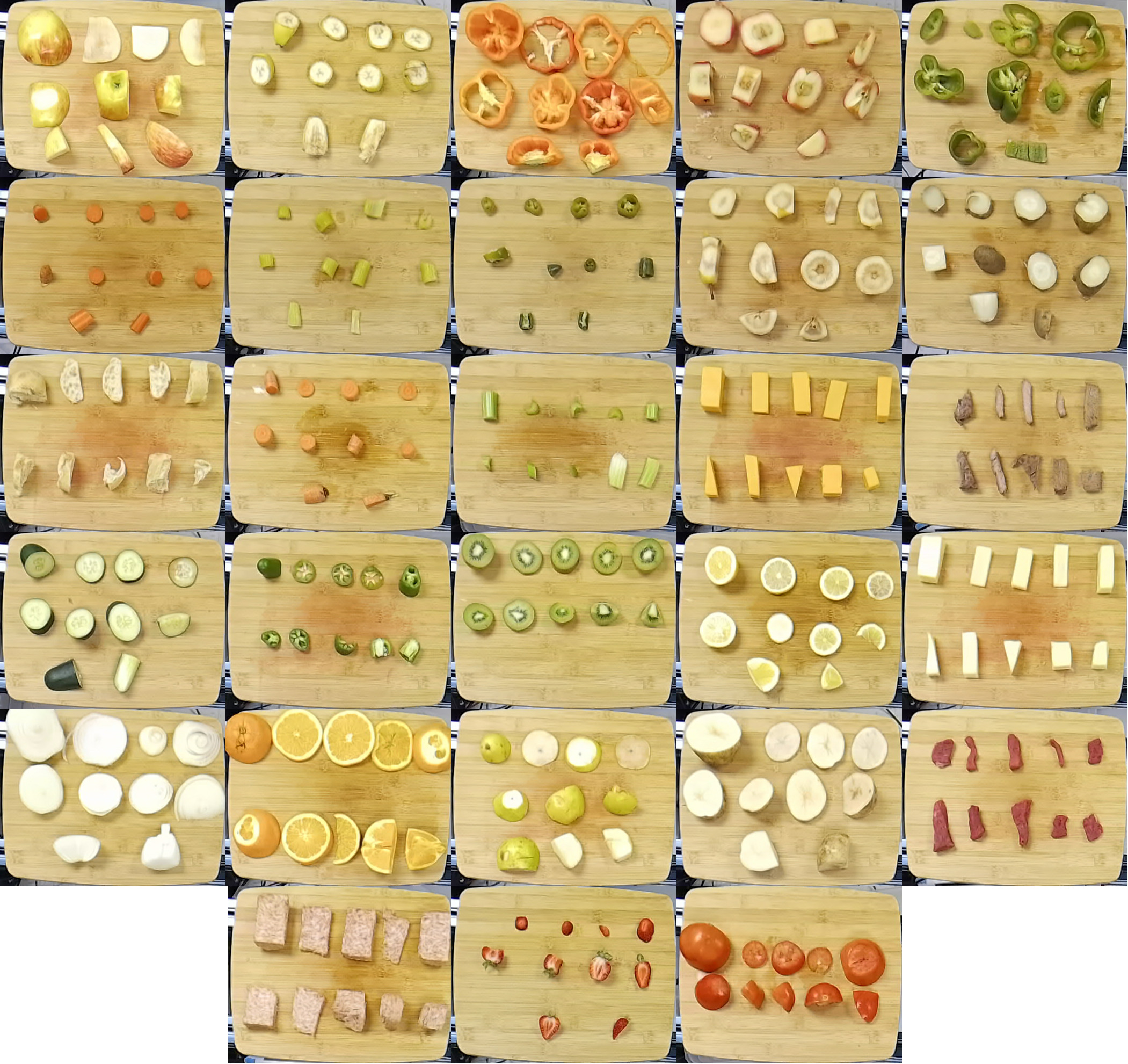}
    \caption{Food item slices. From left to right, top to bottom: apple, banana, bell pepper, boiled apple, boiled bell pepper, boiled carrot, boiled celery, boiled jalapeno, boiled pear, boiled potato, bread, carrot, celery, cheddar, cooked steak, cucumber, jalapeno, kiwi, lemon, mozzarella, onion, orange, pear, potato, raw steak, spam, strawberry, and tomato.}
    \label{fig:food_collage}
    \vspace{-1em}
\end{figure}

Using our robot cutting setup mentioned in Section~\ref{sec:exp_setup}, we taught the robot simple cutting skills using Dynamic Movement Primitives (DMPs)~\cite{kroemer2016meta,schaaldmp}.
Through ridge regression, we fitted DMP parameters to trajectories collected using kinesthetic human demonstrations as in \cite{zhang2019leveraging}. 
Afterwards, we chained DMPs into multiple slicing motions until the specified food item was cut completely through. 

In total, we cut 10 slices each from 21 different food types which include: apples, bananas, bell peppers, bread, carrots, celery, cheddar, cooked steak, cucumbers, jalapenos, kiwis, lemons, mozzarella, onions, oranges, pears, potatoes, raw steak, spam, strawberries, and tomatoes.
The slices are enumerated from 1 to 14 and were created using similar skill parameters across the food types.
The skill parameters vary in slice thickness from 3mm to 50mm, angles from $\pm$30 degrees, and slice orientation where we had normal vs. in-hand cuts when the knife robot cut between the tongs.
There are more than 10 slice types because food items are shaped differently and not all cuts can be executed on every food item, especially the angled cuts.
While the robot is cutting the food items, we collect audio, image, force, and proprioceptive data. 
The resulting slices from various food items are shown in Fig.~\ref{fig:food_collage}.

After the 10 different types of slices have been created, we transfer them to the playing robot setup and begin the data collection process.
We run 5 trials on each of the 10 slices in order to capture variations in the objects' behaviors due to differing initial orientations, positions, and changes over time.
In each trial, we capture a RGBD image from the overhead Kinect and specify the center of the object. 
Afterwards, the robot closes its fingers and pushes down on the object until 10N of force is measured. 
We record the robot's position and forces during this action.
Next, the robot resets to a known position, grasps the object, and releases it from a height of 15cm.
During the push, grasp, and release actions, we record videos from the Realsense and audio from the two contact microphones as mentioned in Section~\ref{sec:exp_setup}.
Videos are only recorded from the FingerVision cameras during the grasp and release actions. 
Additionally, we record the gripper width when the grasp action has finished and save RGBD images from the overhead Kinect before and after each trial.

Our full dataset is available for download \href{https://tinyurl.com/playing-with-food-dataset}{here}\footnote{\url{https://tinyurl.com/playing-with-food-dataset}}. 
The data is located in the appropriately named folders and are sorted first by food type, then slice type, and finally trial number. 
Additionally, we provide food segmentation masks in the silhouette data folder where we used Deep Extreme Cut (DEXTR)~\cite{Man+18} to obtain hand labeled masks of the objects in the overhead and side view images.
Then we fine-tuned a PSPNet~\cite{zhao2017pyramid} pre-trained on the Ade20k~\cite{zhou2017scene} dataset with our manually labeled masks to generate additional neural network labeled segmentation masks.
Finally, we have additional playing data in the old playing data folder, but those slices were all hand cut, and the data was collected in a different environment.

\subsection{Data Processing}
\label{data_pro}
To train the embedding networks, we first extract features from the data. 
We transform the raw audio data from both the cutting data and playing data (during the release action, push-down action, and grasp action) into Mel-frequency cepstrum coefficient (MFCC) features~\cite{logan2000mel} using Librosa~\cite{mcfee2015librosa}.
These features have been shown to effectively differentiate between materials and contact events~\cite{zhang2019leveraging}. 
Subsequently, we use PCA to extract a lower-dimensional representation of the cutting (\cutaudio) and playing (\playaudio) audio features.

To form proprioceptive features (\proprio), we use the robot poses and forces from the push down and grasp actions.
More specifically, for the push down action we extract the final $z$ position (\finalZ) of the robot's end-effector once 10N of force has been reached.
Using this value, we also find the change in $z$ position between the point of first contact and \finalZ \ as \deltaZ, which is an indication of the object's stiffness. 
We retrieve the final gripper width (\gripper) during the grasping action once 60N of force has been reached. These three values are combined to form \proprio.
Labels for each data sample, such as food class label (\foodlabel) and slice type label (\slicelabel), are also created according to the food type and slice type performed during cutting.

\section{Learning Food Embeddings}
\label{approach}

\begin{figure}[t]
    \vspace{-2.5em}
    \centerline{\includegraphics[width=0.9\textwidth]{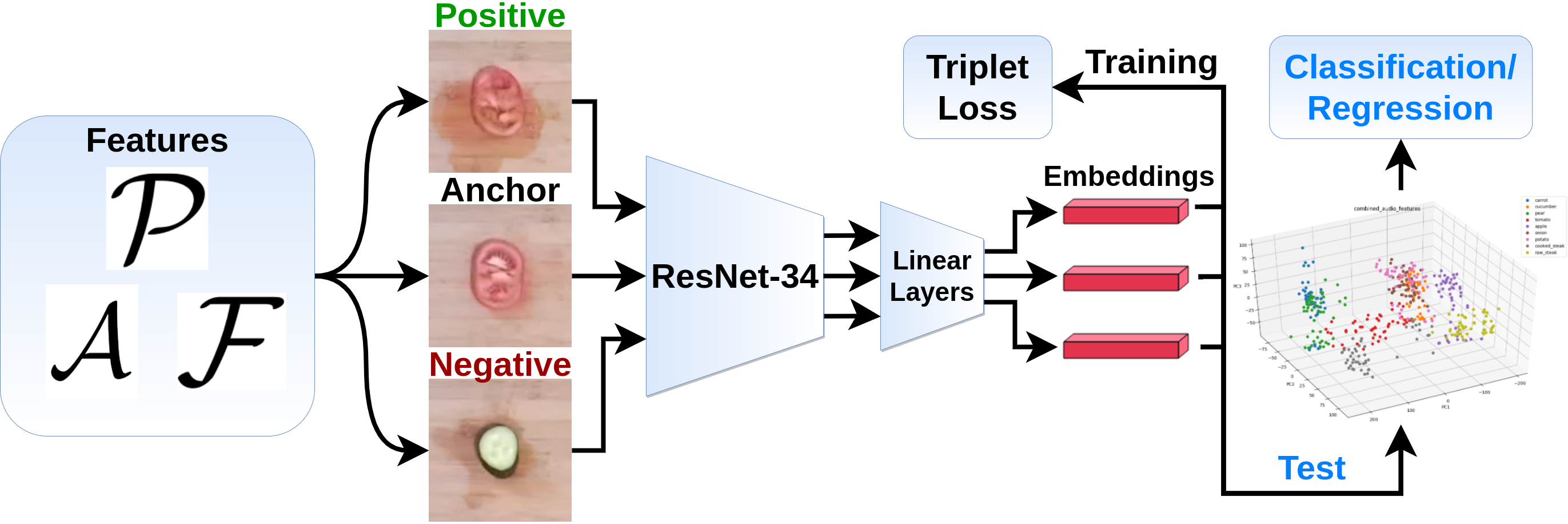}}
    \caption{An overview of our approach. The different features (modalities) defined in Section~\ref{data_pro} are used to form triplets to learn embeddings in an unsupervised manner, which are used for supervised classification and regression tasks (blue text).}
    \label{fig:network}
    \vspace{-1.5em}
\end{figure}

We train convolutional neural networks, in an unsupervised manner, to output embeddings from overhead images. 
Our architecture is comprised of ResNet34~\cite{resnet}, which is pretrained on ImageNet~\cite{imagenet} and has the last fully connected layer removed.
We add an additional three hidden layers, with ReLU activation for the first two, to reduce the dimensionality of the embeddings.
We use a triplet loss~\cite{facenet} to train the network, so similarities across food types are captured in the embeddings.

The different modalities of data mentioned in Section~\ref{data_pro} are used as metrics to form the triplets. 
More specifically, the food class labels (\foodlabel), slice type labels (\slicelabel), playing audio features (\playaudio), cutting audio features (\cutaudio), proprioceptive features (\proprio), and combined audio and proprioceptive features (\playaudio +\proprio) are used as metrics. 
For each sample in the training set, we define the $n$ nearest samples in the PCA feature space (using the L2 norm) as the possible positive samples in a triplet and all other samples as possible negative samples, where $n$ is a hyperparameter ($n = 10$ was used here). 
At training time, triplets are randomly formed using these positive/negative identifiers. 
Fig.~\ref{fig:network} shows an overview of our approach.

To evaluate the usefulness of these learned embeddings, we train multiple 3-layer multilayer perceptron classifiers and regressors for a variety of tasks, using the learned embeddings as inputs. 
These results are presented in Section~\ref{results}. 
When combining multiple modalities, we concatenate the embeddings output from each separate network. 

\section{Experiments}
\label{results}

\begin{table}[h]
\vspace{-2.5em}
\begin{center}
\begin{tabular}{|M{2cm}|M{1.9cm}|M{1.9cm}|M{1.9cm}|M{1.9cm}|M{1.75cm}|}
\hline 
\textbf{Embeddings} & \textbf{Food Type Accuracy - 21 classes (\%)} & \textbf{Hardness Accuracy - 3 classes (\%)} & \textbf{Juiciness Accuracy - 3 classes (\%)} & \textbf{Slice Type Accuracy - 14 classes (\%)} & \textbf{Slice Width RMSE (mm)}\tabularnewline
\hline 
$\mathcal{F}$ & 92.0 & 40.7 & 36.6 & 12.9 & 10.9\tabularnewline
\hline 
$\mathcal{S}$ & 17.1 & 37.0 & 34.9 & \textbf{40.5} & 11.8 \tabularnewline
\hline 
$\mathcal{A}_{play}$ & 85.7 & 35.0 & \textbf{46.0} & 17.1 & 9.9 \tabularnewline
\hline 
$\mathcal{A}_{cut}$ & 93.5 & 33.5 & 45.6 & 16.8 & 11.3 \tabularnewline
\hline 
$\mathcal{P}$ & 49.5 & \textbf{47.1} & 37.0 & 20.0 & \textbf{7.9} \tabularnewline
\hline 
$\mathcal{A}_{play}$+$\mathcal{P}$ & 83.8 & 36.4 & 40.2 & 21.4 & 9.5 \tabularnewline
\hline 
ResNet & \textbf{98.9} & 34.9 & 36.5 & 30.0 & 13.9 \tabularnewline
\hline 
Classifier w/ $\mathcal{A}_{play}$ as input & 84.4 & 40.8 & 34.0 & 30.1 & 34.4 \tabularnewline
\hline 
\end{tabular}
\end{center}
\caption{Baseline and multi-layer perceptron results on 5 evaluation tasks using different learned embedding networks that were trained on our full dataset.}
\label{table:results}
\vspace{-2em}
\end{table}

As mentioned in the previous section, embedding networks were trained using the food class label (\foodlabel), slice type label (\slicelabel), playing audio features (\playaudio), cutting audio features (\cutaudio), proprioceptive features (\proprio), and combined audio and proprioceptive features (\playaudio +\proprio) for creating triplets. 
These embeddings were then used to train the multi-layer perceptrons, mentioned in Section~\ref{approach}, to predict the labels or values for five different tasks: classifying food type (21 classes), predicting slice width (the width of the gripper after grasping), classifying the hardness (3 human-labeled classes - hard, medium, and soft), classifying juiciness (3 human-labeled classes - juicy, medium, dry), and classifying slice type (14 different classes based on the type of cuts the cutting robot performed to generate the slice). 
For our two baselines, we trained convolutional neural networks, with a ResNet34 architecture, that use only visual data and another set of 3-layer mutli-layer perceptrons that use only \playaudio data as input to generate predictions for each of the tasks.

To assess the generalizability of our approach, we evaluated the hardness and juiciness classification tasks based on leave-one-out classification, where we left an entire food class out of the training set and evaluated the trained classifiers on this class at test time. 
We then averaged the results across the 21 leave-one-out classification trials. 
The performance of all the trained networks on each of the tasks are shown in Table~\ref{table:results}. 

As illustrated in Table~\ref{table:results}, the purely visual baseline outperforms our embeddings in the food type classification task due to the vast number of labeled images in the ImageNet dataset that ResNet was pre-trained on. 
However, ResNet performs worse on the other 4 tasks as ImageNet did not contain prior relevant information on the physical properties that are important for these tasks. 
Additionally, ResNet was trained to differentiate object classes instead of finding similarities between classes, so when an entire food category was left out of the training dataset, it most likely had no way of extrapolating the correct answer from previous data.
Meanwhile, our embeddings contained auxiliary information that encoded the similarity of slices through various multimodal features, without ever being given explicit human labels.
This indicates that our interactive multimodal embeddings provided the neural networks with a greater ability to generalize to unseen data as compared to the supervised, non-interactive baselines. 

Additionally, the results show that the audio embeddings provide some implicit information that can help the robot distinguish vegetable types. 
On the other hand, it makes sense that the proprioceptive embeddings are more useful at predicting hardness and slice width as their triplets were generated using similar information.
However, absolute labels were never provided when training the embedding networks, so the learned embeddings encoded this relative information themselves.
It should also be noted that in the hardness and juiciness leave-one-out classification tasks, some food types, such as tomatoes, were more difficult to classify when left out of the training dataset than others, such as carrots. 
This may be due to the small size of our diverse dataset, which has few items with similar properties.

Finally, with respect to the slice type prediction task, there were poor performances across the board due to the inherent difficulty of the task.
Due to the variability of shapes between food items, the resulting slices generated by the cutting robot, while executing the same actions, differed greatly at times.
Thus, it was to be expected that only the embeddings trained using slice type labels performed relatively well on this classification task.
Overall, the results of the evaluations above show that certain embeddings performed better on relevant tasks, which supports the hypothesis that they are encoding information on different material properties and can be applied in different use cases.

\begin{table}[t]
\begin{center}
\begin{tabular}{|M{2cm}|M{2cm}|M{2cm}|M{2cm}|}
\hline 
\multirow{2}{*}{\textbf{Embeddings}} & \textbf{Hardness} & \textbf{Juiciness} & \textbf{Cooked}\tabularnewline
 & \textbf{Accuracy (\%)} & \textbf{Accuracy (\%)} & \textbf{Accuracy (\%)}\tabularnewline
\hline 
$\mathcal{A}_{play}$ & 98.0 & 62.9 & 98.9\tabularnewline
\hline 
$\mathcal{P}$ & 63.0 & 68.4 & 60.6\tabularnewline
\hline 
$\mathcal{A}_{play}$+$\mathcal{P}$ & \textbf{99.7} & \textbf{70.6} & \textbf{99.1}\tabularnewline
\hline 
ResNet & 90.5 & 66.1 & 90.4\tabularnewline
\hline 
Classifier w/ $\mathcal{A}_{play}$ as input & 82.1 & 67.4 & 88.8\tabularnewline
\hline 
\end{tabular}
\end{center}
\caption{Results on 3 evaluation tasks for different learned embedding networks that were trained using the auxiliary cooked vs. uncooked dataset.}
\vspace{-2em}
\label{table:results_boiled}
\end{table}

\vspace{-1em}
\subsubsection{Auxiliary Study with Additional Cooked vs. Uncooked Food Data:}

As an addendum to the 21-food class dataset described in Section \ref{data_col}, we collected an additional dataset of boiled food classes to further explore and evaluate our method's ability to detect whether a food item is cooked or not through interactive learning. 
The additional boiled food classes collected were: apples, bell peppers, carrots, celery, jalapenos, pears, and potatoes. 
Each item was boiled for 10 minutes.
Note that for these additional cooked food classes, we did not have the robot cut the food slices due to difficulties grasping the objects. 
We combined these boiled classes with their uncooked counterparts from the full dataset to form a 14-class dataset and conducted a subset of evaluations on the embeddings learned from this dataset, shown in Table \ref{table:results_boiled}.

The auxiliary study with the boiled food dataset shows that the playing audio data is effective at autonomously distinguishing between cooked and uncooked food items without being provided any human-provided labels. 
This is likely because cooking food significantly changes its material properties. The high performance of the (\playaudio + \proprio) embeddings on the hardness, juiciness, and cooked leave-one-out classification tasks on this smaller dataset demonstrates the ability of our approach to generalize to new data when a food class with similar properties was present during training (apples and pears, potatoes and carrots, bell peppers and jalapenos).

\begin{figure}[t]
    \begin{subfigure}{0.48\textwidth}
        \centering
        \includegraphics[width=\textwidth]{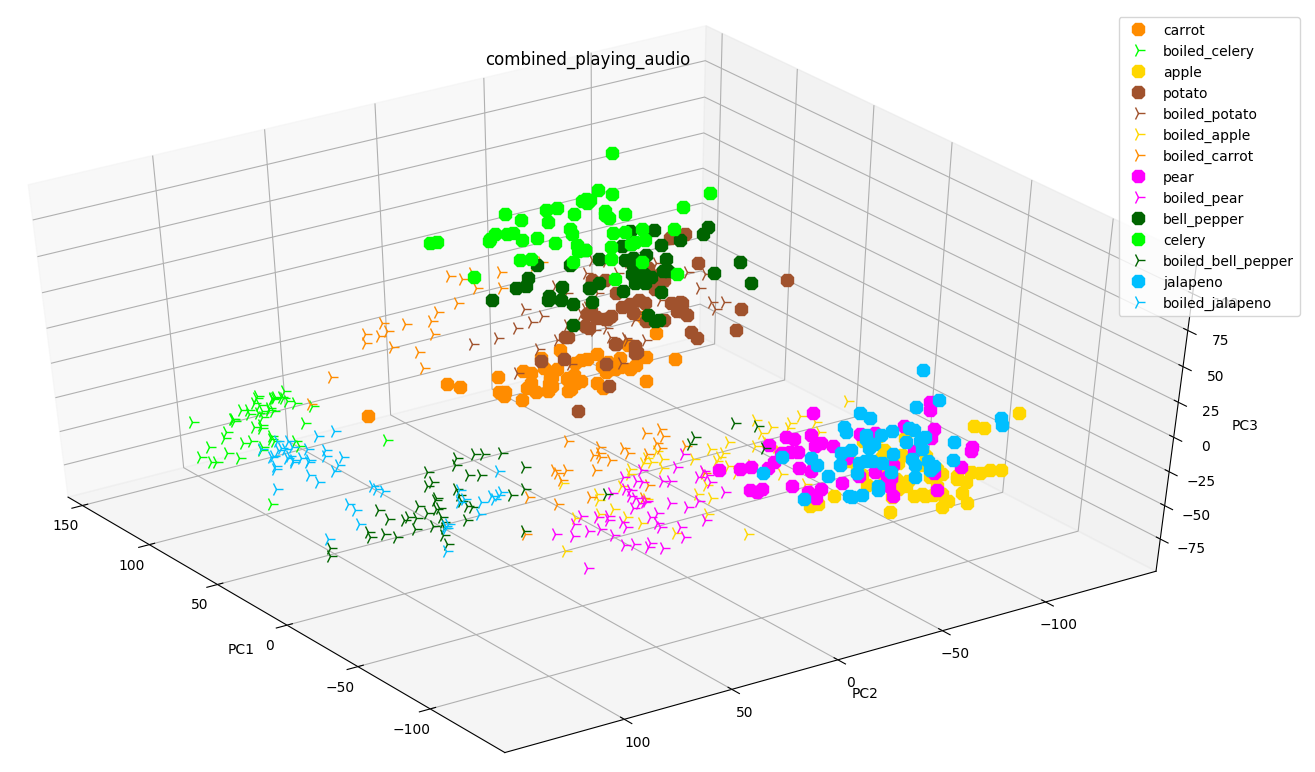}
        \caption{Playing audio PCA features.}
        \label{fig:combined_audio}
    \end{subfigure}
    \begin{subfigure}{0.48\textwidth}
        \centering
        \includegraphics[width=\textwidth]{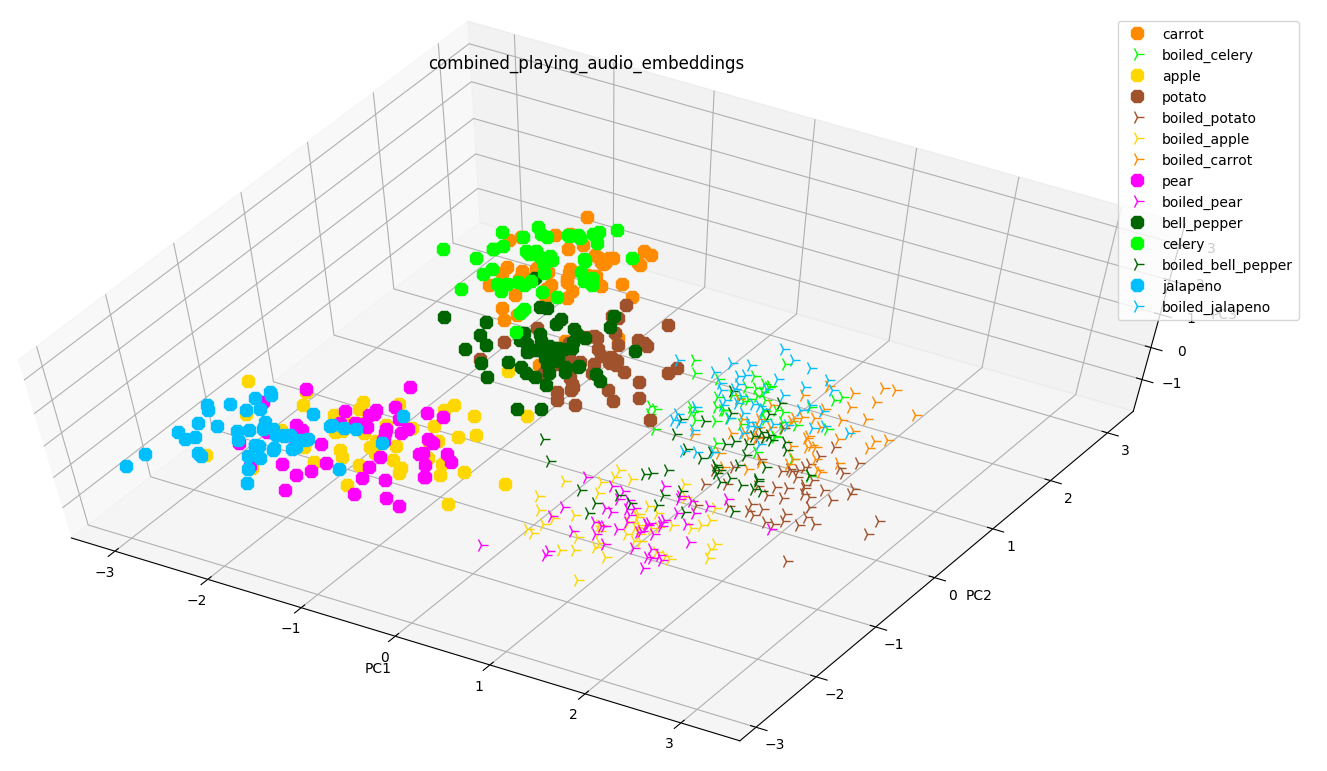}
        \caption{Playing audio learned embeddings.}
        \label{fig:combined_audio_embeddings}
    \end{subfigure}
    \caption{Fig.~\ref{fig:combined_audio} visualizes playing audio features using PCA. 
    Fig.~\ref{fig:combined_audio_embeddings} visualizes the embeddings learned using the playing audio features also in PCA space.
    }
    \vspace{-1em}
\end{figure}

Fig.~\ref{fig:combined_audio} shows the playing audio features (\playaudio) of the different cooked and uncooked food classes in this dataset using the top 3 principal components. Fig.~\ref{fig:combined_audio_embeddings} visualizes the learned embeddings based on \playaudio, also in PCA space. 
As shown in the plots, there is a distinct separation between the boiled and raw foods in the audio feature space and also in the learned embedding space. 
Within the cooked and uncooked groupings, there are certain food types that tend to cluster together. 
For example, the uncooked pear and apple cluster close together, which makes sense given the similarity of these two fruits. 
Interestingly, they remain clustered close to one another even after they are cooked, even though there is a shift in the feature space between cooked and uncooked.



\section{Conclusions and Future Work}

In this work, we have presented a novel dataset consisting of autonomously collected audio, proprioceptive, force, and visual data that was recorded while a robot played with a variety of slices from 21 unique food types that have different shapes and properties.
In addition, we learned visual embedding networks that utilized our multimodal dataset to encode properties of food items using a triplet loss formulation.
These learned embeddings were shown to encode similarities between food types without explicit human labeling and outperformed normal visual-only and audio-only baselines on a variety of tasks.
We hope others can utilize our publicly available dataset to explore novel ways of deciphering and encoding the material properties of food items in order to further propel research on food robotics forward.

For example, an interesting extension to this work would be to apply state of the art computer vision tracking algorithms in order to observe the deformations and movements of the slices using the videos captured from the RealSense and FingerVision cameras.
This could serve as an additional metric for creating triplets or possibly be used as a basis for creating dynamics models for different types of food.
In addition, since we recorded videos when cutting the slices, it may be possible to predict the shape of resulting slices given an action or vice versa.

For us, the next challenges we hope to tackle using this dataset include: monitoring the progression of food as it is being cooked in order to inform the robot when to intervene, and using the learned embeddings to better execute manipulation tasks such as cutting, flipping, mixing, picking, and placing food items.
If we collect data that is complementary to this dataset then we will append the additional data to the original dataset. 
We also welcome others to contribute as well. 
Overall, we believe that this work is an exciting step towards autonomously learning about deformable food items through robotic interactions and play.

\bibliographystyle{abbrv}
\bibliography{references}

\end{document}